\documentclass[sigconf]{acmart}
\usepackage{xcolor}
\usepackage{float}
\usepackage{caption}
\usepackage{subcaption}
\usepackage{url}
\usepackage{enumitem}
\usepackage{multirow}
\usepackage{verbatim} 
\usepackage{subfiles}
\usepackage{wrapfig, lipsum, booktabs}
\usepackage{mathtools}
\usepackage{makecell}
\usepackage{amsmath}
\usepackage{soul}
\usepackage{graphicx}
\usepackage{pifont}

\usepackage[compact]{titlesec}
\setcopyright{acmlicensed}
\copyrightyear{2018}
\acmYear{2018}
\acmDOI{XXXXXXX.XXXXXXX}
\acmISBN{978-1-4503-XXXX-X/18/06}

\begin{document}

\title{SARN: Structurally-Aware Recurrent Network for Spatio-Temporal Disaggregation}

\author{Bin Han}
\email{bh193@uw.edu}
\affiliation{%
  \institution{University of Washington}
  \city{Seattle}
  \country{USA}
}

\author{Bill Howe}
\email{billhowe@uw.edu}
\affiliation{%
  \institution{University of Washington}
  \city{Seattle}
  \country{USA}
}

\begin{abstract}
    Open data is frequently released spatially aggregated, usually to comply with privacy policies.  But coarse, heterogeneous aggregations complicate learning and integration for downstream AI/ML systems.  In this work, we consider models to disaggregate spatio-temporal data from a low-resolution, irregular partition (e.g., census tract) to a high-resolution, irregular partition (e.g., city block). We propose an overarching model named the Structurally-Aware Recurrent Network (SARN), which integrates structurally-aware spatial attention (SASA) layers into the Gated Recurrent Unit (GRU) model. The spatial attention layers capture spatial interactions among regions, while the gated recurrent module captures the temporal dependencies. Each SASA layer calculates both global and structural attention --- global attention facilitates comprehensive interactions between different geographic levels, while structural attention leverages the containment relationship between different geographic levels (e.g., a city block being wholly contained within a census tract) to ensure coherent and consistent results. For scenarios with limited historical training data, we explore transfer learning and show that a model pre-trained on one city variable can be fine-tuned for another city variable using only a few hundred samples. Evaluating these techniques on two mobility datasets, we find that on both datasets, SARN significantly outperforms other neural models (5\% and 1\%) and typical heuristic methods (40\% and 14\%), enabling us to generate realistic, high-quality fine-grained data for downstream applications.
\end{abstract}



\maketitle

\section{Introduction}\label{sec:intro}
High-quality, longitudinal, and freely available urban data, coupled with advances in machine learning, improve our understanding and management of urban environments \cite{han2023adapting}. Over the last two decades, cities have increasingly released datasets publicly on the web, proactively, in response to transparency regulation. For example, in the US, all 50 states and the District of Columbia have passed some version of the federal Freedom of Information (FOI) Act. While this first wave of open data was driven by FOI laws and made national government data available primarily to journalists, lawyers, and activists, a second wave of open data, enabled by the advent of open source and web 2.0 technologies, was characterized by an attempt to make data ``open by default" to civic technologists, government agencies, and corporations~\cite{verhulst2020emergence}.

Open data affords study of urban dynamics~\cite{vogel2011understanding, yu2017spatiotemporal}, but, with few exceptions~\cite{zhang-aggregated-obs,singh2021}, models assume access to individual-level data, while privacy concerns motivate the release of aggregated data~\cite{Green2017OpenDP} such that aggregated data tends to be the norm in some fields~\cite{burrell2004, lozano2009, law2018}. Although aggregation can protect individual privacy~\cite{privacy_healthcare, privacy_metering, privacy_sensor} (without strong guarantees~\cite{privacy_no_gaurantee_1, privacy_no_gaurantee_2}), it complicates integration of multiple datasets (e.g., aligning tract-level to block-level data) and destroys information useful for training. For example, demographers have studied disaggregation techniques to uncover local hotspots erased by tract-level aggregation\cite{Monteiro2018}. The New York City (NYC) taxi open datasets serve as another notable example, widely used in spatio-temporal traffic forecasting literature. Much of the work has relied on NYC taxi data collected before June 2016, when individual-level records were accessible \cite{mo2022cross, zhang2023mcl, jin2022selective, han2023adapting}. Subsequently, taxi records have been released at aggregated taxi zone levels, forcing applications to rely on out-of-date pre-2016 data.

Motivated by these issues, we study the spatio-temporal disaggregation problem --- to break down the aggregated urban data from a coarse, irregular spatial partitioning to a finer, irregular partitioning. The disaggregation problem has been studied in different domains. Spatial disaggregation methods in the demography literature tend to rely on heuristics: areal weighting, which assumes all variables are constant per unit area~\cite{Comber2019SpatialIU}, and dasymetric mapping~\cite{Comber2019SpatialIU}, which assumes sufficient auxiliary data is available to predict the target variable using linear methods.  Disaggregation has also been considered in the machine learning literature, outside of the spatiotemporal context, where the goal is to learn individual-level models given group-level aggregated data~\cite{zhang-aggregated-obs,linear_disaggregatioin}. However, these approaches explicitly seek a model where the target variable at the individual level is determined \textit{only} by its parent group's features as opposed to learning from global patterns across groups.  In a spatiotemporal context, any region may influence any other: traffic on the bridge can increase wait times in midtown, for instance.  
\begin{figure}[htp]
    \centering
    \includegraphics[width=0.48\textwidth, height=6cm]{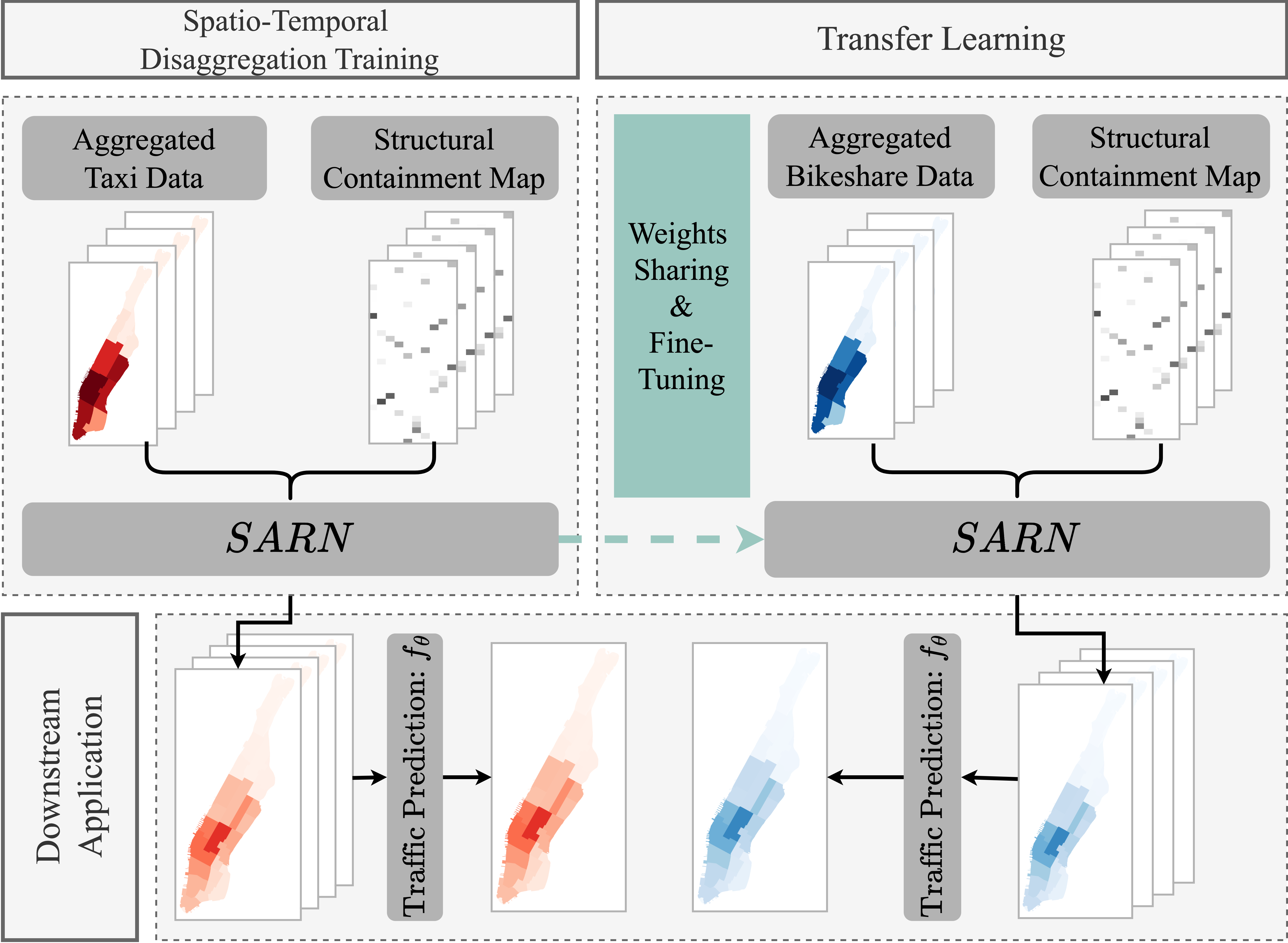}
    \caption{Disaggregation problem setting and its utilities. Our proposed Structurally-Aware Recurrent Network (\textbf{SARN}) is trained with coarse, aggregated data and structural containment maps. The trained model can be utilized in downstream applications, such as traffic prediction. It also can be adopted to disaggregate another variable through weight-sharing and fine-tuning with only a limited amount of data.}
    \label{fig:teaser}
    \vspace{-0.25cm}
\end{figure}

In this paper, we evaluate neural architectures for the disaggregation problem, and demonstrate the utility of the disaggregated data (See Figure \ref{fig:teaser}). We proposed an overarching model, Structurally-Aware Recurrent Network (SARN), which integrates structurally-aware spatial attention layers (SASA) into the Gated Recurrent Unit (GRU) model. The SASA layers capture convoluted spatial interactions among regions, while the gated recurrent module captures the temporal dependencies. Specifically, each SASA layer calculates two types of attention --- global and structural. Global attention attends each coarse region to all fine regions, facilitating a comprehensive, macro-level geographic interactions. To enhance learning efficiency and effectiveness, we leverage the structural property exhibiting between geographic levels, \textit{containment} --- each fine level region is entirely contained in a coarse level region. Such phenomenon is practically seen in political boundaries. Exploiting this structural property, we create binary containment maps to be used in structural attention calculation, promoting the generation of coherent and consistent values across varying geographic levels. We conduct three experiments targeting various disaggregation levels with two mobility datasets from NYC. Our results demonstrate that SARN significantly enhances disaggregation performance and converges faster than other neural models.

In addition to our primary disaggregation findings, we conduct three supplementary experiments to enhance the practicality of the disaggregation task and demonstrate the utility of our model:  1) \textit{Transfer Learning}. We relax the assumption that temporal historical data is always available and explore pre-training on variable $A$, then fine-tuning on variable $B$. Our results demonstrate the remarkable effectiveness of transfer learning, yielding competitive outcomes compared with specialized models, with only a minimal number of samples. 2) \textit{Downstream Application}. The disaggregated data can be valuable for downstream applications. We use traffic demand prediction as an example, as accurate traffic demand prediction is essential for cities to allocate transportation resources effectively, especially in fine regions (e.g., city blocks), potentially reducing traffic congestion. We train several models with the disaggregated taxi data and compare their performance to those trained with ground truth data. Our results show remarkable authenticity of the disaggregated data compared to the real data. 3) \textit{Individual Points Synthesis}. We quantitatively show that our disaggregated values have a similar distribution to the ground truth values over a set of regions, via mutual information and Kullback–Leibler divergence statistics. Then we qualitatively demonstrate that a simple sampling scheme can be used to generate a realistic individual-level of points. 

Overall, we make the following contributions:
\begin{itemize}[leftmargin=0.14in]
    \setlength{\itemsep}{2pt}
    \item We propose the Structurally-Aware Recurrent Network (SARN), designed to disaggregate spatio-temporal data. The structurally-aware spatial attention (SASA) layers model spatial interactions, while the gated recurrent module captures temporal dependencies. Within each SASA layer, global attention is calculated to enable macro-level interactions between levels. By employing containment maps, structural attention is calculated to produce coherent and consistent values between different levels.
    
    \item We test three disaggregation resolutions and two mobility datasets. On NYC taxi data, SARN achieves a 5\% improvement over the best neural models and a 40.1\% improvement over the best classical methods on average. For NYC bikeshare data, the average improvements are 1.2\% and 13.6\%, respectively. Furthermore, SARN converges faster than other neural models.
    
    \item We consider a transfer learning scenario, acknowledging that sufficient training data may not always be available to train an effective neural model for every city variable. Our results demonstrate that, within the same city, parameters learned from one variable can be used to make predictions on another variable through rapid fine-tuning (1-2 hours versus 12 hours of pre-training), achieving competitive performance.
    
    \item We demonstrate the utility of disaggregated data in downstream traffic demand prediction. The prediction difference between models trained with disaggregated values and ground truth values is trivial (0.78\%), indicating that the disaggregated data is of high-quality and closely matches the ground truth values. Additionally, we synthesize individual-level data via sampling and show that synthesized points have similar distributions to the true values, evidenced by high mutual information (0.736) and low KL-divergence (0.058) statistics.
    
\end{itemize}
Our work is available at this repository for reproducibility \footnote{\url{https://github.com/BeanHam/2023-urban-disaggregation}}.

\section{Related Work}\label{sec:literature}
\subsection{Spatial Disaggregation}
Spatial disaggregation is frequently used in demography to infer individual records from spatially aggregated regions such as census data~\cite{Qiu2019GeospatialDO, Wardropa2018SpatiallyDP}, frequently applied in population analysis (e.g., dynamic population mapping). Dynamic data on human population distribution at high resolution is important for a wide spectrum of activities and real-life applications, such as urban applications of business locations, transportation planning, city service management etc. \cite{zong2019deepdpm} Other applications include disease mapping \cite{Arambepola2020ASS} and vaccination coverage \cite{Utazi2018ASR}. 

A popular heuristic is areal weighting (AW), which assumes the value per unit area is constant~\cite{Goodchild1993AFF}.  This method is common in practice due to its simplicity, general reliability, and ability to computed from a single variable. Pycnophylactic interpolation (PI), first introduced by Tobler in 1979 \cite{pycnophylactic}, works similarly to areal weighting, but with a refinement to avoid discontinuities at the boundaries. PI distributes the values from source regions to target regions iteratively , considering a weighted average of its nearest neighbors to avoid discontinuities while preserving the total count. 

Dasymetric mapping/interpolation/weighting (DM) approach is a family of methods that use the relationships between multiple variables to improve the estimated distribution. Unlike areal weighting, dasymetric mapping uses auxiliary information to generate a weighted distribution of source values onto the target regions. Recent methods incorporate machine learning approaches to model the relationships between auxiliary data and source values and determine the weights for disaggregation \cite{Qiu2019GeospatialDO, Monteiro2018, Monteiro2019}. Sources of auxiliary information include satellite images \cite{stevens2015}, 3D building information \cite{Qiu2019GeospatialDO}, mobile phone usage data \cite{Deville2014}, terrain elevation and human settlement \cite{Monteiro2019}, and other variables more broadly available at high spatial and temporal resolution. Despite the popularity of dasymetric mapping methods in spatial disaggregation, there are several potential limitations of them \cite{Comber2019SpatialIU}. They are wholly dependent on auxiliary variables, which are not always available (e.g., remote sensing images or land type data in population disaggregation tasks). Additionally, the performance of the model depends on the correlation between auxiliary data and source values, which may be low due to lack of relevant datasets or noisy measurements. 

Deep learning methods have also been directly applied to the disaggregation problem in recent years. For example, Zong et al. \cite{zong2019deepdpm} proposed the DeepDPM model, which uses a super-resolution CNN to analyze both spatial and temporal patterns using coarse data and point-of-interest information. In the urban computing domain, several recent studies have focused on generating fine-grained traffic flow from coarse-grained data using super-resolution techniques. Ouyang et al. \cite{ouyang2020fine} introduced the UrbanFM model, which employs a feature extraction module and a distributional upsampling module to infer fine-grained crowd flow from coarse observations. Liu et al. \cite{liu2023road} proposed the RATFM model, which leverages prior knowledge of road networks to learn the road-aware spatial distribution of fine-grained traffic flow. Both studies use super-resolution techniques, but their problem setting differs slightly from ours. Their coarse observations are at the pixel level, while our coarse values are at predefined geopolitical region levels (e.g., tracts, blocks). In an urban image, each region may cover numerous pixels that share the same value.

\subsection{Learning From Aggregated Observations}
In addition to the spatial disaggregation methods, it is also possible to learn individual level probabilistic models from aggregated data, without first decomposing aggregated data into individual instances. The problem is generally referred to as \textit{learning from aggregated observations}, where supervision labels are given to sets of instances instead of individual instances, while the goal is still to make inferences on unseen individuals \cite{zhang-aggregated-obs, singh2021, law2018}. Zhang et al. \cite{zhang-aggregated-obs} present a general probabilistic framework capable of accommodating a variety of aggregate observations for various tasks, such as classification and regression. They show that simple maximum likelihood solutions can be applied to various differentiable models such as deep neural networks and gradient boosting machines. 
Ma et al. \cite{ma2020learning} studied the problem of learning nonlinear dynamics from aggregated data, where individual trajectory data is not available. They proposed a partial differential equation to describe the density evolution of aggregated data, which is subsequently combined with a Wasserstein generative adversarial network (WGAN) in the training process. 
Wei et al. \cite{wei2023universal} studies a more specific problem --- classification from aggregate observations (CFAO), where the supervision is provided to groups of instances, instead of individual instances. They presented a novel universal method of CFAO, which generates unbiased estimators of the classification risk for arbitrary losses. 

\subsection{Spatio-Temporal Traffic Prediction} 
Spatiotemporal disaggregation can be cast as a prediction problem, for which neural methods have received significant attention. Traffic forecasting is a spatiotemporal prediction problem that is a building block for a number of mobility applications\cite{jin2023transferable}, such as trip planning (e.g., travel time estimation \cite{li2018multi}, route representation\cite{liu2023unified}) and resource optimization (e.g., dynamic fleet management \cite{zheng2022supply}). Earlier work~\cite{li2017diffusion, li2021spatial, yu2017deep} used graph convolutional network (GCNs) to capture spatial interactions and sequential models such as Long-Short Term Memory (LSTMs) or Gated Recurrent Unit (GRUs), to model temporal dependencies. Later work \cite{zhu2021ast, zheng2020gman, lan2022dstagnn} utilized spatial and/or temporal attention mechanisms to capture more complicated spatial and temporal dynamics and correlations. Recent work applies spatio-temporal transformers\cite{yan2021learning, liu2023spatio, fang2022learning}. Spatio-temporal transformers have also been applied in other areas, such as visual tracking \cite{yan2021learning}, trajectory prediction\cite{chen2021s2tnet}, and human action recognition\cite{ahn2023star}.

\section{Datasets}\label{sec:data}
\begin{figure}[htp]
    \centering
    \vspace{-0.25cm}
    \includegraphics[width=0.46\textwidth]{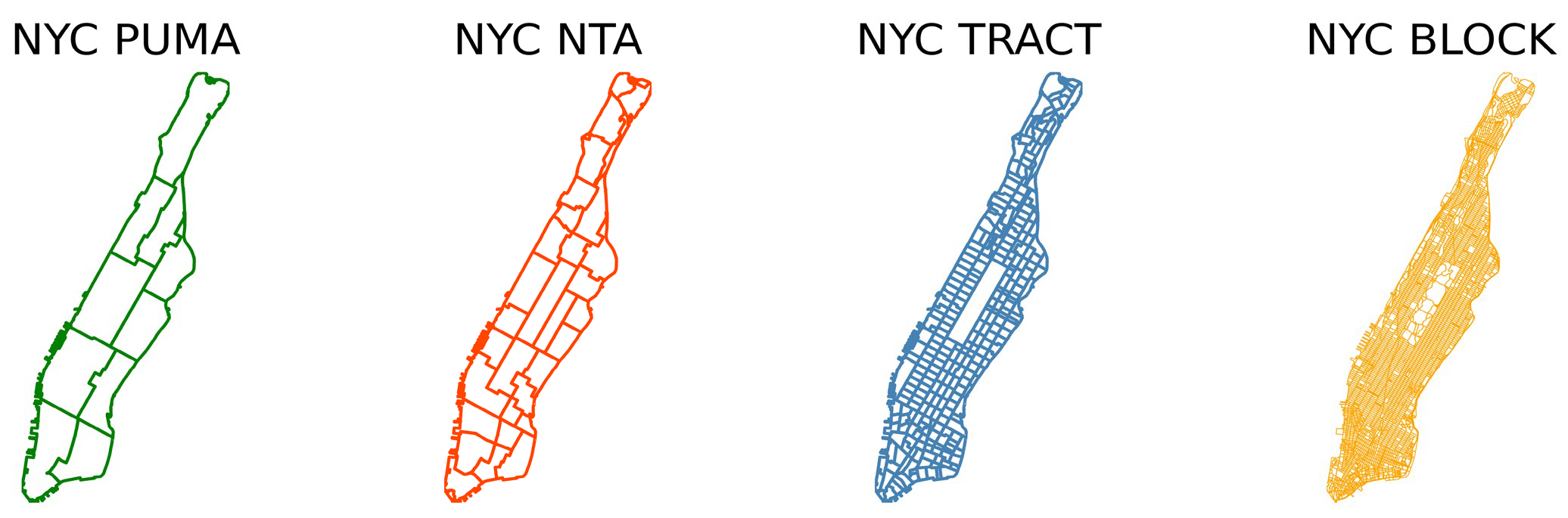}
    \vspace{-0.25cm}
    \caption{Four distinct geographic aggregation levels for the same NYC area. PUMA regions are the largest and most coarse. BLOCK regions are smallest and finest.}
    \label{fig:geo-visualization}
    \vspace{-0.25cm}
\end{figure}

In our research, we use two mobility datasets --- NYC taxi data and NYC bikeshare data. Each dataset contains individual-level records. 
\begin{itemize}[leftmargin=0.14in]
    \item \textbf{NYC Taxi Data}: NYC taxi trip data is collected from NYC Open Data portal\footnote{\url{https://opendata.cityofnewyork.us/data/}} from 01/01/2016 to 06/30/2016. The raw data are presented in tabular format. Each record summarizes the information for one single taxi trip, which contains the longitude and latitude of the location of the trip start.
    
    \item \textbf{NYC Bikeshare Data}: NYC bikeshare data is collected from NYC DOT \footnote{\url{https://citibikenyc.com/system-datafrom}} 01/01/2021 to 12/31/2021. Similar to the taxi data, the raw data are presented in tabular format. Each data point represents one bike trip, including the longitude and latitude of the location where the bike was unlocked. The dataset spans the entire year, allowing the inspection of seasonal variations. 
\end{itemize}
These datasets represent two examples of mobility events in the same city (taxi and bikeshare in NYC). This collection enables study of variance of our method (same city, same application domain, but different variable). Moreover, NYC taxi and bikeshare datasets afford transfer learning studies (e.g., training on taxi data and evaluating on bikeshare data) to address the situation where available data may be limited. For NYC, we have restricted our experiments to the Manhattan region. We use four pre-defined geographic levels offering varying levels of spatial resolution: Public Use Microdata Areas (PUMA), Neighborhood Tabulation Areas (NTA), Census Tracts (TRACT), and Census Blocks (BLOCK) (see Figure \ref{fig:geo-visualization}). We obtain the 2010 boundaries for these levels from the NYC open data portal. Figure \ref{fig:geo-visualization} shows the resolutions of four geographic levels. 
\begin{table}[htp]
    \small
    \tabcolsep=0.35cm
    \renewcommand*{\arraystretch}{1}
    \centering
    \begin{tabular}{|c|c|c|c|c|}
        \hline
        \textbf{Data} & \textbf{PUMA} & \textbf{NTA} &\textbf{TRACT} & \textbf{BLOCK} \\ \hline
        \multirow{2}{*}{\thead{NYC \\ Taxi}} & 1443.01 & 450.94 & 51.17 & 3.86 \\ & (1776.03) & (689.16) & (73.32) & (8.03) \\ \hline
        \multirow{2}{*}{\thead{NYC \\Bikeshare}} & 189.24 & 59.14 & 6.71 & 0.50 \\ & (279.28) & (97.86) & (13.09) & (2.51)\\ \hline 
    \end{tabular}
    \caption{Descriptive statistics of two datasets. The numbers are the average hourly taxi/bike counts per geographic region. The standard deviations are in the parenthesis.}
    \label{tab:descriptive-statistics}
    \vspace{-0.75cm}
\end{table}

\subsection{Data Processing}\label{sec:data_processing}
Given a dataset of individual events $(lat, lon, time)$, we construct raster representations (for CNN models) and vector (for other neural models) representations of spatiotemporal event data as follows:

\paragraph{\textbf{Vector}} We aggregate the event counts by hour and region for each of the four geographic levels. For example, there are 10 Public Use Microdata Areas (PUMA); a single hour of data is represented as simply a 10-element vector.  The order of elements is arbitrary. More generally, the dataset is represented as $X \in \mathbb{R}^{N \times d}$, where $N$ is the total number of hours, and $d$ is the number of regions at a given geographic level (e.g., $d$ = 10 for PUMA; $d$=32 for NTA).  

\paragraph{\textbf{Raster}} We first define a 128$\times$256  grid and plot the $N$ regions at each geographic level. Each region has a corresponding value from the vector input --- $X_{n}, n\in\{1,2,\dots,N\}$, and the number of pixels located within the region --- $C_{n}, n\in\{1,2,\dots,N\}$. Then each pixel in each region takes the value of $\frac{X_{n}}{C_{n}}$.  Pixels shared by more than one region are assigned to whichever region is smaller to avoid ambiguity. The cost is that we lose information during translation since the raster representation does not represent region boundaries precisely, especially at BLOCK level.

We compute descriptive statistics, including means and standard deviations, for the counts of each dataset at different geographic levels. These statistics are shown in Table \ref{tab:descriptive-statistics}. For NYC Taxi data, the standard deviations are large demonstrating the significant variability in the count values between regions. For instance, Inwood in upper Manhattan has considerably fewer taxi rides compared to the more central areas of Manhattan.  In the case of the NYC Bikeshare, we notice significantly smaller average values at the TRACT and BLOCK levels. This observation can be attributed to the distribution of the bikes, which may vary in frequency across different regions within the city. As a result, the average values of these datasets are lower at the more granular levels. 

\section{Models}\label{sec:model}
\setlength{\belowdisplayskip}{2pt} \setlength{\belowdisplayshortskip}{2pt}
\setlength{\abovedisplayskip}{2pt} \setlength{\abovedisplayshortskip}{2pt}

\begin{figure*}[ht]
    \centering
    \includegraphics[width=\textwidth]{visualizations/sarn.png}
    \caption{Model architecture of our proposed \textbf{SARN}. (a) Original architecture of the GRU model. Fully-connected layers (FC) are applied on input $X^i$. (b) Our proposed \textbf{SARN} architecture. The fully-connected layers are replaced with our proposed structurally-aware spatial attention (SASA) layers. (c) Our proposed SASA layer. A SASA layer is composed with multiple spatial attention heads. Each spatial attention head calculates global and structural attention across regions from two different geographic levels. The containment map is incorporated to reflect the real-world structural hierarchy between geographic levels, and ensures the generation of coherent values across different resolutions.}
    \label{fig:sarn}
    \vspace{-0.25cm}
\end{figure*}

In \S\ref{sec:notation} we formally formulate the disaggregation problem. In \S\ref{sec:containment}, we introduce the structural property and containment maps that connect different levels of the aggregation hierarchy. In \S\ref{sec:sasa}, we detail the architecture of the structurally-aware spatial attention (SASA) layer, including the calculation of global and structural attention using the containment maps. In \S\ref{sec:sarn}, we present our proposed SARN model, which integrates SASA layers into the GRU layer. The model architecture is visualized in Figure \ref{fig:sarn}

\subsection{Notations \& Problem Formulation}\label{sec:notation}
We adopt the following notation to be used throughout the paper:
\begin{itemize}[leftmargin=0.14in]
    \setlength{\itemsep}{2pt}
    \item $g^i$ for $i \in \{0,1,\dots,G\}$ denotes a geographic aggregation level. The levels are labeled in ascending order in terms of number of regions, i.e., the lowest resolution is assigned 0 and the highest resolution is assigned $G$.  In our experiments, the input is always the lowest resolution, e.g. PUMA.  Therefore, it is assigned with the level $g^0$. Level $g^G$ has the largest number of regions, which is BLOCK. Level $(g^1,g^2,\dots,g^{G-1})$ are intermediate levels.

    \item $|g^i|$ denotes the number of regions at geographic level $g^i$. $|g^G|$ denotes the number of regions at the BLOCK level, which we use to normalize the loss terms.  That is, the loss for all levels is expressed in average error per city block to afford direct comparisons across levels.
    
    \item $X^i \in \mathbb{R}^{N \times |g^i|}$ denotes the ground truth ${N \times |g^i|}$ vector at geographic level $g^i$ assuming $N$ total hours. We include a subscript $(t)\in [1,N]$ to indicate a single timestep $X^i_t \in \mathbb{R}^{|g^i|}$. To indicate a range of timesteps $t$ to $t+T$, we will write $X^i_{t:t+T}$.

    \item $\hat{X}^i \in \mathbb{R}^{N \times |g^i|}$ denotes the predicted data at geographic level $g^i$ assuming $N$ total hours. We will indicate slices of this vector using the same subscript notation as with ground truth. 
\end{itemize}
The disaggregation problem is defined as --- given a time series of traffic values in the past $T$ time steps at coarse level $g^i$, we aim to learn a function $f$ that disaggregates the data to a finer level $g^j$: 
$$f([X^i_{t-(T-1)}, \dots, X^i_t];\theta) \rightarrow [X^j_{t-(T-1)}, \dots, X^j_t], \forall j>i$$

\subsection{Structural Property --- Containment}\label{sec:containment}

\begin{figure}[htp]
    \centering
    \vspace{-0.25cm}
    \includegraphics[width=0.48\textwidth]{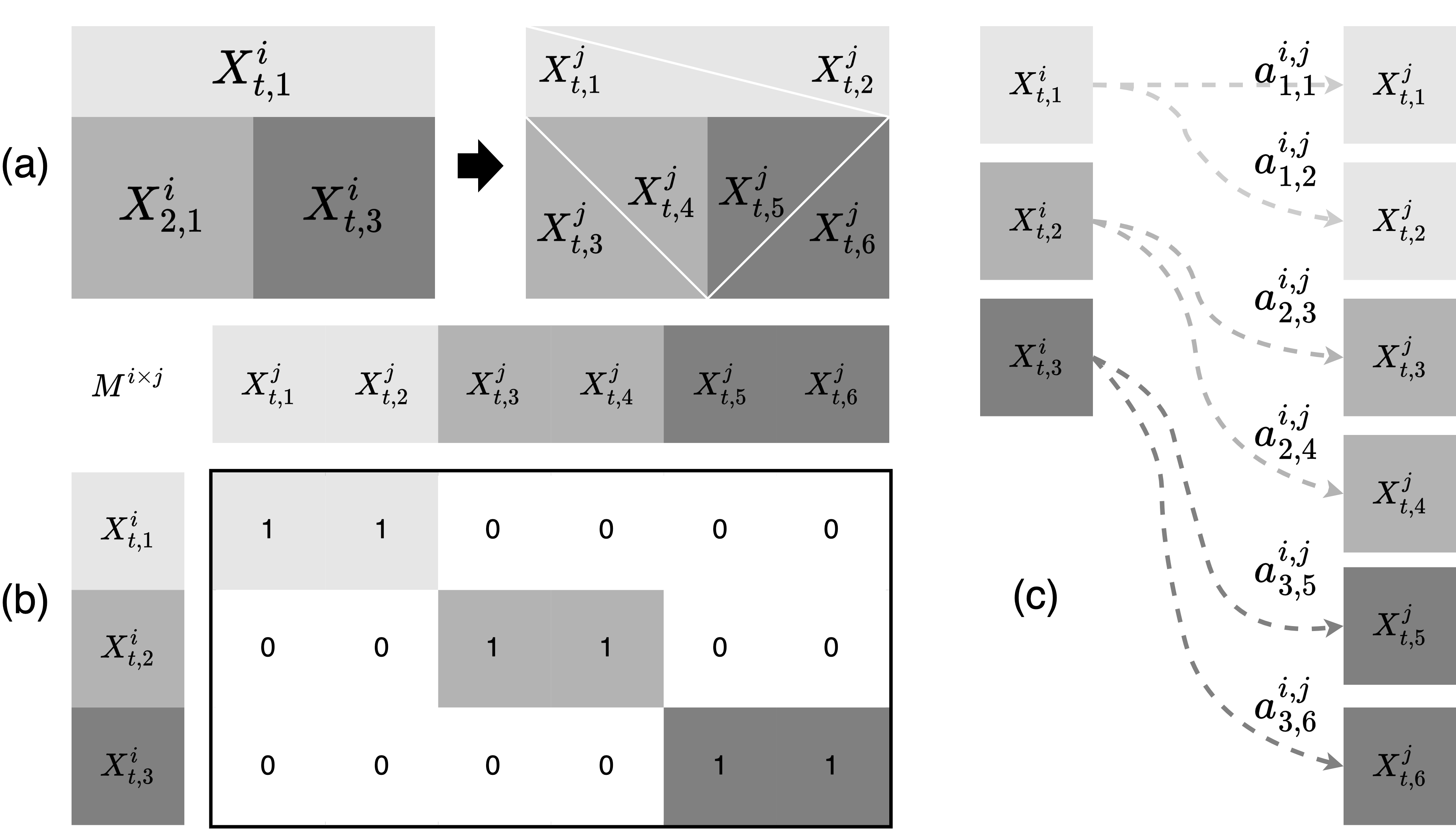}
    \caption{(a) Containment relationships between geographic level $g^i$ and $g^j$. Each region from $g^i$ is wholly contained in a region from $g^j$. (b) Binary containment map $M^{i,j}$ between georaphic level $g^i$ and $g^j$. (c) Structural-aware spatial attention is calculated among connected regions from two levels.}
    \label{fig:containment}
    \vspace{-0.25cm}
\end{figure}

The geographic levels we use exhibit structural problem \textit{containment} --- each region at level $g^i, i>1$ is wholly contained in a region at level $g^{i-1}$ such that regions are hierarchical. For example, one census tract comprises multiple census blocks. Assuming containment, we can identify a one-to-many map among the regions from any two geographic levels. Let $M^{i,j}, \forall j>i$ be a $|g^i|\times|g^j|$ binary matrix representing the containment relationship between geographic levels $g^i$ and $g^j$. A value of 1 at position $M^{i,j}[c,r]$ indicates that region $r$ at level $g^j$ is contained in region $c$ at level $g^i$. Figure \ref{fig:containment}(a) shows an example where three regions from level $g^i$ at a single $t$th timestamp are split into six regions from level $g^j$. Figure \ref{fig:containment}(b) presents the corresponding binary containment map.

\subsection{Structurally-Aware Spatial Attention Layer}\label{sec:sasa}

Spatio-temporal transformers have been applied to urban prediction tasks (Section \ref{sec:literature}). The multi-head spatial attention layers in the transformer blocks can capture complex spatial relationships among different regions. In our disaggregation task, we also leverage the attention mechanism to extract spatial information. The spatial attention is calculated across two geographic levels with different spatial dimensions. We propose a structurally-aware spatial attention (SASA) layer (see Figure \ref{fig:sarn}(c)), which consists of multiple structurally-aware spatial attention heads ($SASA_{head}$). Each head performs two attention calculations --- a global attention and a structural attention. Given input $X^i_t \in \mathbb{R}^{1\times |g^i|}$ from geographic level $g^i$, the global attention score is calculated as:
\begin{equation}
    Att_{g} = Softmax(Q^TK/\sqrt{|g^j|}) \in \mathbb{R}^{|g^i|\times |g^j|} 
\end{equation}
\begin{equation}
    Q = X^i_tW_Q \in \mathbb{R}^{1\times |g_i|}, W_Q \in \mathbb{R}^{|g_i|\times |g_i|}
\end{equation}
\begin{equation}
    K = X^i_tW_K \in \mathbb{R}^{1\times |g_j|}, W_K \in \mathbb{R}^{|g_i|\times |g_j|}
\end{equation}
The global attention enables each coarse region to attend to all finer regions, facilitating comprehensive interactions across different geographic levels. This macro-level interaction allows the model to capture broader dependencies. The structural attention (See Figure \ref{fig:containment}(c)) uses the \textit{containment} map to force the attention to happen only among the connected regions: 
\begin{equation}
    Att_{s} = Softmax(M^{i,j} \odot Q^TK/\sqrt{|g^j|}) \in \mathbb{R}^{|g^i|\times |g^j|}     
\end{equation}
The containment maps impose a constraint that requires the values of child regions to sum up to the value of their parent region. This ensures that the predicted values remain coherent and consistent across different geographic resolutions. By enforcing this hierarchical relationship, the model maintains the integrity of the spatial dependencies, allowing for accurate disaggregation results. The global and structural attention outputs are added and go through a feed-forward projection layer:
\begin{equation}
    H = FNN(V \cdot (Att_{g}+Att_{s}))\in \mathbb{R}^{1\times |g^j|}
\end{equation}
\begin{equation}
    V = X^i_nW_V \in \mathbb{R}^{1\times |g_i|}, W_V \in \mathbb{R}^{|g_i|\times |g_i|}
\end{equation}
A structurally-aware spatial attention layer (SASA) is composed with multiple spatial attention heads, enhancing the model capacity:
\begin{equation}\label{spatial_attention_equation}
    SASA = FNN(Concat(H_1, H_2, \dots, H_N))
\end{equation}
$$H_i = SASA_{head, i}(Q, K,V)$$
The same calculation works with temporal inputs. Given $T$ hours of input $X^i_{t:t+T}$, we reshape the dimension $X^i_{t:t+T} \in \mathbb{R}^{T \times 1 \times |g^i|}$ and calculate spatial attention based on Eq (\ref{spatial_attention_equation}). In the regular attention calculation [44], a residual connection layer is applied after the multi-head attention layer: $X_t^i + MultiHead(Q,K,V)$. In the disaggregation task, the input and output have different spatial dimensions. Therefore, residual connection is not applicable. 

\subsection{Structurally-Aware Recurrent Network}\label{sec:sarn}

The spatial attention layer only captures spatial dependencies across different geographic levels. The temporal information, which is also critical for spatio-temporal prediction tasks, is not represented except as a set of 
independent samples. Recent research has increasingly employed spatio-temporal transformers for spatio-temporal traffic forecasting. The temporal attention layers capture the temporal dependencies in the input sequences. Temporal attention layers are typically preferred to conventional recurrent models due to their ability to learn long-distance dependencies and their efficient implementations in parallel settings. Unlike recurrent models, where information encoded at each time step is retained solely for the subsequent step and quickly diminishes in influence, attention layers calculate attention scores across the entire sequence concurrently, thereby promoting shared sequential information. However, GRUs are still widely used in spatio-temporal prediction tasks \cite{lu2022spatio, chen2023traffic, ren2023transformer, wang2020traffic}. Additionally, our setting involves relatively short temporal lengths, and we empirically find that GRUs outperform temporal transformer-based models (see \S\ref{sec:results}). 

The comparison of SARN and the original GRU model is visualized in Figure \ref{fig:sarn}(a) and (b). Let $X^i_{t:t+T}$ be $T$ hours of input from geographic level $g^i$. Then for each element in the input sequence, an original GRU layer computes:
$$r_t = \sigma(X^i_tW_{xr} + h_{t-1}W_{hr}+b_r)$$
$$z_t = \sigma(X^i_tW_{xz} + h_{t-1}W_{hz}+b_z)$$
$$n_t = tanh(X^i_tW_{xn} + (r_t \odot h_{t-1}W_{hn} + b_n)$$
$$h_t = z_t \odot h_{t-1} + (1-z_t) \odot n_t$$
where $X^i_t \in \mathbb{R}^{1\times |g^i|}$ is the input at time $t$, $h_{t-1}$ is the hidden state of the layer at time $t-1$ or the initial hidden state at time 0. $h_t$ is the new hidden state, which is the hidden input for next time.

In the GRU model, $X_t^iW_{xr}, X_t^iW_{xz}$ and $X_t^iW_{xn}$ changes the spatial dimension from coarse level $|g^i|$ to finer level $|g^j|$. In our SARN model, we replace the three matrix multiplication with our SASA layers as introduced in \S\ref{sec:sasa}. Then SARN is capable of modeling spatial and temporal dependencies concurrently (See Figure \ref{fig:sarn}(b)). The calculation for the three gates becomes:
$$r_t = \sigma(SASA_{xr} + h_{t-1}W_{hr}+b_r)$$
$$z_t = \sigma(SASA_{xz} + h_{t-1}W_{hz}+b_z)$$
$$n_t = tanh(SASA_{xn} + (r_t \odot h_{t-1}W_{hn} + b_n)$$
$$h_t = z_t \odot h_{t-1} + (1-z_t) \odot n_t$$
$SASA_{xr}, SASA_{xz}$ and $SASA_{xn}$ are three separately trained spatial attention layers for the three gates.

\subsection{Loss}\label{sec:loss}
We calculate $\ell_1$ loss at the target disaggregation level. However, since $\ell_1$ loss divides by the size of the vector, meaning that the $\ell_1$ loss will have different units at each aggregation level.  For example, the $\ell_1$ loss at the TRACT level is in units of ``events per TRACT region'' while the $\ell_1$ loss at the BLOCK level is in units of ``events per BLOCK region'', leaving the two losses incomparable. Therefore, to meaningfully compare the losses at different target disaggregation level, we normalize all loss terms to be events-per-region at a selected aggregation level.  In our experiments, we normalize to the BLOCK level to assist with interpretation: we consider taxi rides per city block as more familiar than taxi rides per PUMA region, for example. We use the following loss at the target level $g^i$: 
\begin{equation}\label{eq:regular_loss}
    \ell_1^{i} = \frac{1}{|g^G|}\sum|\hat{X}^i_{n} - X^{i}_{n}|
\end{equation}

\section{Experiments}\label{sec:experiments}
In this section, we describe the baseline models to compare against for the disaggregation task. Additionally, we describe the setup for the supplementary studies of downstream application, transfer learning and point synthesis. 

\subsection{Baseline Models}\label{sec:baselines}
As our baseline, we test several heuristic-based disaggregation methods used commonly in the literature and in practice, as well as more advanced neural approaches:
\begin{itemize}[leftmargin=0.14in]
    \item {\textbf{Constant Weighting (CW)}: The event count for a region in the output is assumed to be $\frac{1}{N}$ of the count in the containing region in the input, where $N$ is the number of output regions contained in the input region.  That is, this heuristic assumes the events are uniformly distributed, regardless of size and shape of the region.}

    \item {\textbf{Areal Weighting (AW)}: The event count for an output region $r$ contained in an input region $c$ is assumed to be proportional to the fractional area of $r$ in $c$.  That is, $|r| = \frac{A(r)|c|}{A(c)}$, where $A(r)$ is the area of region $r$ and $|c|$ is the event count in region $c$.  That is, this heuristic assumes events are uniform per unit area.}

    \item \textbf{Historical Ratios (HR)}: The event count of an output region $r$ is assumed to be a fixed proportion of the count of an input region $c$. The fixed proportion is computed as the average proportion from all historical timesteps.  

    \item \textbf{Feedforward Neural Network (FNN)}: The event count for an output region $r$ is assumed to be a (non-linear) function of all input regions, not just the containing region. The function is modeled as a multi-layer feedforward neural network trained on historical data. This approach attempts to capture the interdependencies between regions. We use several fully-connected intermediate layers. 

    \item \textbf{Convolutional Neural Network (CNN)}: The event count at each output region $r$ is extracted from the raster representation at the output aggregation level (e.g., city blocks).  That raster representation in turn is learned from the raster representation of the input aggregation level (e.g., PUMA) using a typical UNet architecture\cite{unet}.  This approach models the non-linear relationships between regions, while emphasizing local effects through convolutional layers. 

    \item \textbf{Gated Recurrent Unit (GRU)}: Where FNN only models the spatial connections (by assuming that each timestep is an independent smaple), GRU can capture temporal patterns. We chose T=5 as the temporal sequence length across all our experiments based on the study \cite{han2023adapting}.  Otherwise, this model is similar in inputs and outputs to FNN. We use several intermediate GRU layers. 

    \item \textbf{UrbanFM\cite{ouyang2020fine}}: a disaggregation model that generates fine traffic flow data from coarse, aggregated data using super-resolution technique. It adopts grid representations of traffic flow, which is similar to the CNN baseline. We modified the input PUMA image to be $64\times 32$\footnote{As mentioned in \S\ref{sec:literature}, their coarse observations are at pixel level, while our coarse values are at predifined geopolitical regions. When we process rasterized image inputs, pixels lying in the same region share the same value. Given this preprocessing, the $N_2$ normalization and distribution recovery layer can no longer be applied in the original UrbanFM model, which might negatively impact the results.}.     
    
    \item \textbf{Spatio-Temporal Transformer (ST-Transformer)}: a model consisting a spatial and a temporal transformer. The spatial transformer shares the same spatial attention layer described in \S\ref{sec:sasa}, with global attention only. The temporal transformer follows the set up in \cite{liu2023spatio} --- given the output from the spatial transformer $\hat{X}^i_{t:t+T}\in \mathbb{R}^{T\times |g^j| \times 1}$, we add hour-of-day and day-of-week index to the featurs, and generate a feature embedding. Then regular multi-head self-attention mechanism is applied. .   
\end{itemize}
We test three disaggregation tasks --- from PUMA to NTA, TRACT and BLOCK level respectively. Batch size, learning rate, early stop tolerance are set to 32, 1e-4, and 50 respectively. For STT model, the feature embedding size, number of temporal attention head, and number of temporal transformer layer are set to 64, 4, and 4 respectively. For our SARN, we only use one GRU layer, and each SASA layer uses 4 $SASA_{head}$. We use 1xNVIDIA T4 GPU, 16 GB RAM, x4 vCPUs, for each task.


\subsection{Downstream Application}\label{sec:application}

To demonstrate the high quality and utility of the disaggregated data, we use traffic demand prediction as the downstream application task. Traffic demand prediction involves forecasting the number of vehicles required in a region within a specific time frame. Accurate traffic demand prediction enables efficient allocation of transportation resources, reducing mismatches between supply and demand, alleviating traffic congestion, and enhancing mobility service experiences. Coarse-level predictions (e.g., PUMA) are less useful as they tend to miss fine-grained human mobility patterns. Therefore, using disaggregated values at finer resolutions is more beneficial. The specific task is to predict the next hour's taxi demand given the previous twelve hours of data. We first disaggregate taxi values from PUMA to NTA, TRACT, and BLOCK levels using our proposed SARN model on the original validation split. We then train several traffic prediction models (LSTM\cite{lstm}, AGCRN\cite{agcrn}, STSGCN\cite{stsgcn}) with both the disaggregated data and the original ground truth data at each level. Mean Absolute Error (MAE) is reported on the test set to evaluate performance.

\subsection{Transfer Learning}\label{sec:transfer-learning}
As mentioned in \S\ref{sec:intro}, there are urban settings where the availability and quality of fine-grained data are limited for model training, posing a barrier to implement our model. This data availability issue is not just a limitation to our model, but across all neural models that require sufficient training; it is also universal to all urban computing tasks where neural models excel. In these cases, we can consider whether a model pre-trained on one variable can be used to make predictions for another variable.  We can use our models in this transfer learning setting by training on, say, taxi data, then fine-tuning on a limited amount of bikeshare.  This approach can significantly increase the value of open data, by allowing one dataset to be used in different applications.  The idea is reasonable, given the interconnectedness of city dynamics as a complex system~\cite{integrativeUrbanAI}: everything depends on everything else.  Moreover, the main signals being learned by any model are influenced by the built environment, human activity, and population density patterns, all of which tend to be exhibited in any spatio-temporal dataset in the same domain.

\subsection{Event Data Synthesis}\label{sec:point-synthesis}

While the BLOCK aggregation level is high-resolution (smaller than census tracts for example), they remain spatially aggregated and cannot be directly utilized by applications requiring individual point data. Additionally, if we can synthesize individual taxi rides for NYC Taxi data after June 2016 given the aggregated taxi zone data, it would significantly enhance the utility of the datasets. A simple approach to synthesize points is to learn a high-resolution aggregation at BLOCK level, then randomly sample within these regions to generate individual points, considering the physical built-in environment (e.g., building, hotel etc.). If a region has $k$ events, we can sample $k$ events within its boundaries. Quality control is important for the synthesized data. Quantitatively, we can compare the distributions of real and disaggregated values across a set of BLOCK regions. We consider two statistics to measure similarities --- Mutual Information (MI) and KL-divergence. 
MI measures the probabilistic dependency between two variables, where a higher value indicates greater shared information between the variables. KL divergence is a measure of entropy increase due to the use of an approximation distribution to the true distribution. A lower KL divergence value signifies a better match between the true distribution and our approximation. We also assess the distribution similarity qualitatively through visualization.

\section{Results}\label{sec:results}
\begin{table*}[!ht]
    \tabcolsep=0.18cm
    \renewcommand*{\arraystretch}{1.1}
    \centering
    \begin{tabular}{|c|c|c|c|c|c|c|c|c|c||c|}
        \hline
        \textbf{Dataset} & \textbf{Resolutions} & \textbf{CW} & \textbf{AW} & \textbf{HR} & \textbf{UrbanFM} & \textbf{CNN} & \textbf{FNN} & \textbf{GRU} & \textbf{STT} & \textbf{SARN} \\ \hline
        \multirow{3}{*}{NYC Taxi} & PUMA $\rightarrow$ NTA &  1.8295 &  1.4072 &  0.4652 & 0.3989 & 0.3601 &  0.2300 & \underline{0.2255} &  0.2729 & \textbf{0.2073*} (\rotatebox{90}{\color{green!80!black}\ding{225}}8.1\%)  \\  
        & PUMA $\rightarrow$ TRACT &  1.8416 & 2.0651 & 0.9650 & 0.8049 & 0.7324 & 0.5673 & \underline{0.5631} &  0.6473 & \textbf{0.5443*} (\rotatebox{90}{\color{green!80!black}\ding{225}}3.4\%) \\
        & PUMA $\rightarrow$ BLOCK &  3.0947 & 2.9705 & 1.7609 & 1.8576 & 1.5306 & 1.4515 & \underline{1.4278} & 1.4305 & \textbf{1.3763*} (\rotatebox{90}{\color{green!80!black}\ding{225}}3.6\%) \\
        \hline
        \multirow{3}{*}{NYC Bikeshare} & PUMA $\rightarrow$ NTA &  0.1678 & 0.1212 & 0.0619 & 0.0834 & 0.0587 & 0.0497 & \underline{0.0478} & 0.0601 & \textbf{0.0473*}(\rotatebox{90}{\color{green!80!black}\ding{225}}1.1\%) \\
        & PUMA $\rightarrow$ TRACT &  0.2973 & 0.2945 & 0.1650 & 0.1787 & 0.1591 & \underline{0.1497} & 0.1505 & 0.1584 & \textbf{0.1483*}(\rotatebox{90}{\color{green!80!black}\ding{225}}1.5\%) \\
        & PUMA $\rightarrow$ BLOCK &  0.7882 & 0.7574 & 0.2488 & 0.3726 & 0.2481 & 0.2492 & \underline{0.2341} & 0.2513 & \textbf{0.2316*}(\rotatebox{90}{\color{green!80!black}\ding{225}}1.1\%) \\ \hline
    \end{tabular}
    \caption{Disaggregation results from traditional and neural-based methods. The unit of the error is count/(block * hour). \textbf{Bold} and \underline{underline} numbers indicate the best and second result in the corresponding disaggregation task respectively. * indicates statistical significance test from bootstrap sampling. Our \textbf{SARN} has the best performance among five out of six tasks. \rotatebox{90}{\color{green!80!black}\ding{225}} indicates the performance increase compared with the GRU model.}
    \label{tab:disaggregation_results}
    \vspace{-0.5cm}
\end{table*}

In this section, we present our experimental results that answer the following questions:
\begin{enumerate}[leftmargin=0.3in,label=(Q\arabic*)]
    \item Do neural methods improve performance over commonly used heuristic methods? (Yes)  (Section \S\ref{sec:results-neural-methods})    
    \item Does our proposed SARN improve disaggregation performance relative to neural methods?  (Yes) Does SARN converge faster than competing methods? (Yes) (Section \S\ref{sec:results-sarn})
    \item Is the disaggragated data from SARN authentic and of great quality compared to real data? Can it be used in downstream applications? (Yes) (Section \S\ref{sec:result-application})
    \item Can we fine-tune a disaggregation model trained on one variable and have it perform competitively on a different variable? (Yes) (Section \S\ref{sec:result-transfer-learning})
    \item Can disaggregation models synthesize realistic individual events from  aggregated values? (Yes) (Section \S\ref{sec:result-point-synthesis})
\end{enumerate}

\subsection{All Neural Methods Outperform Heuristics}\label{sec:results-neural-methods}
Table \ref{tab:disaggregation_results} presents the errors of the disaggregation task using both heuristic and neural methods. We observe that:
\begin{itemize}[leftmargin=0.12in]
    \item As the target disaggregation resolution increases, the error increases across all methods, as expected. Predicting city block dynamics from large regional aggregates is a difficult problem. 
    \item Popular heuristic-based disaggregation methods, such as CW and AW, are non-competitive across all resolutions and datasets. While the underlying heuristics are reasonable, they cannot well capture the dynamics. The distribution of taxi rides and bikeshare trips follows human activity patterns influenced by temporal effects and non-linear interactions between regions, which simple heuristics cannot access.    
    \item HR significantly outperforms other heuristics (53.5\% average increase over AW), indicating that temporal patterns are generally predictive. However, HR still overlooks the non-linear temporal and spatial variations present in the data, limiting its accuracy.     
    \item Neural-based methods outperform all common heuristics in nearly all disaggregation tasks. GRU models capture additional temporal dependencies, thus outperforming models that solely rely on spatial information. GRU emerges as the best-performing baseline model (2.3\% average increase over FNN).     
    \item UrbanFM is outperformed by its counterpart image model CNN. We contribute the huge performance gap to the difference in setting of two disaggregation problems. The $N_2$ normalization and distribution recovery layer proposed for the original problem can no longer be applied in our setting, thus negatively impacting the results. 
\end{itemize}
Overall, the results show that neural models, especially GRU models, outperform heuristic-based approaches.

\subsection{SARN Improves Neural Results}\label{sec:results-sarn}

From Table \ref{tab:disaggregation_results}, we observe that our proposed SARN has the best performance across all six tasks! With taxi data, it delivers a 5.0\% improvement over the best neural models (GRU) on average and a 40.1\% improvement over the best classical methods (HR) on average. With bikeshare data, we see relatively smaller but still notable improvements of 1.2\% and 13.6\%, respectively. The largest improvement is observed when disaggregating from PUMA to NTA with NYC Taxi data, where SARN reduces the average hourly taxi count error in the entire Manhattan region by 301 compared to the GRU model! For statistical significance tests, we use bootstrap sampling for 10,000 iterations \cite{bootstrap1, bootstrap2}: On each iteration, we sample the predictions, then calculate error and compute the proportion that are better than the baseline to provide an empirical estimate of the p-value. We use 0.05 as the significance level. With both the taxi and bikeshare dataset, SARN consistently outperforms GRU with statistically significant improvement. 

We notice that STT models generally perform poorly on both datasets, even worse than FNN models. The performance gap is noticeably large on simpler tasks, from PUMA to NTA. We conjecture two reasons: 1) STT has overly complicated architectures for simpler disaggregation tasks. STT has over 150K parameters when disaggregating PUMA to NTA, while GRU only has 10K and SRAN has 35K. Models with more parameters do not necessarily enhance performance, but could possibly contain noisy signals that hurt the results. 2) STT shares the same spatial transformer layer as SARN. The only difference is the temporal module. Temporal attention is not effective and is incapable of producing consistent results compared with recurrent modeling unit in the disaggregation task. Although attention-based solutions are increasingly popular due to their success, we find that relatively short sequences and smaller capacity models can still motivate recurrent models.

\begin{figure}[ht]
    \centering
    \includegraphics[width=0.46\textwidth]{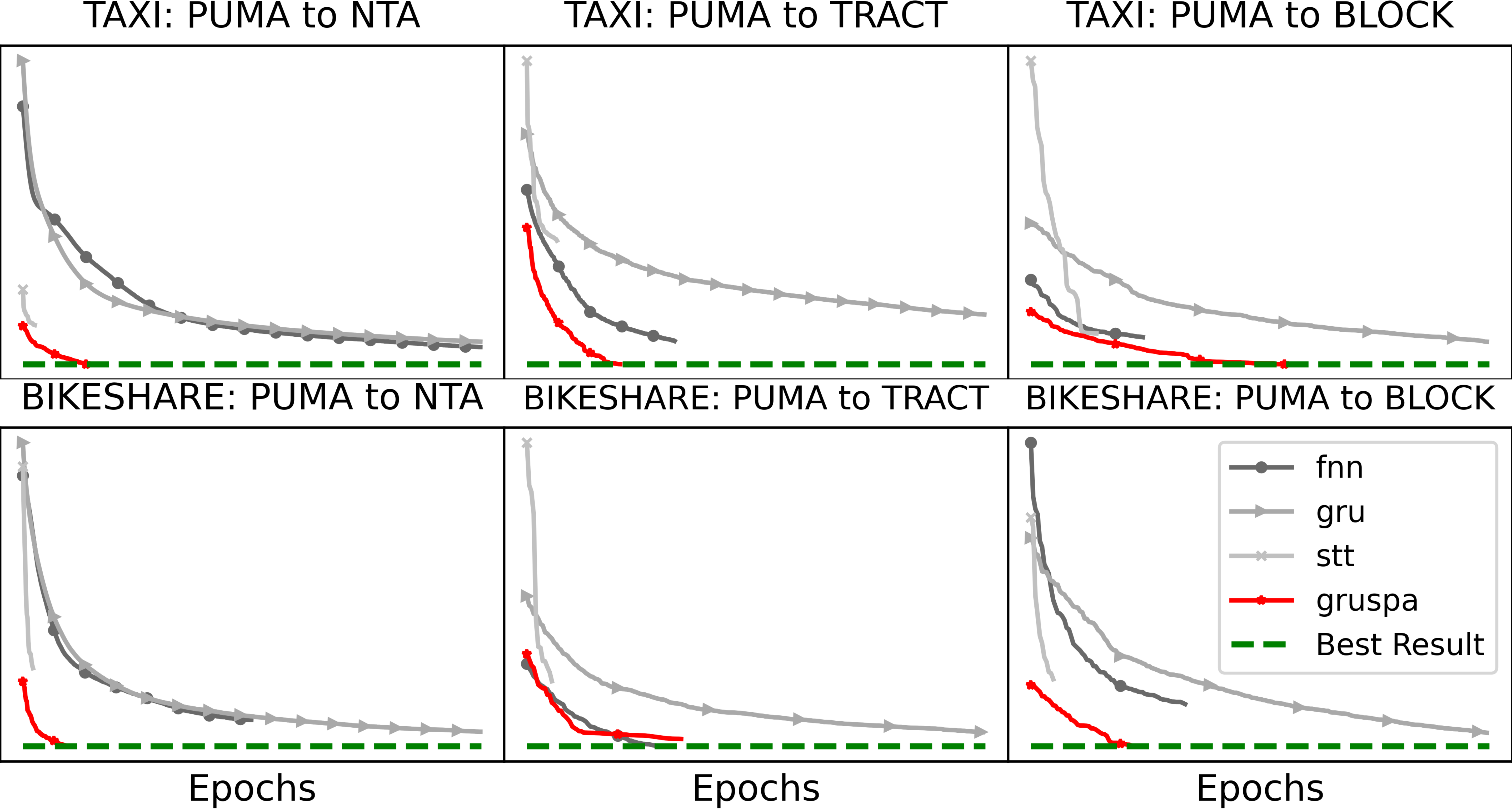}
    \vspace{-0.25cm}
    \caption{Convergence plots of neural models. Top three plots: show NYC Taxi dataset. Bottom three plots: shows NYC Bikeshare dataset. Red line represents our proposed SARN model, and green line indicates the best results. In five out of six tasks, our SARN converged faster than other neural methods.}
    \label{fig:convergence}
    \vspace{-0.25cm}
\end{figure}

\paragraph{\textbf{SARN Converges Faster than Other Neural Model:}} Figure \ref{fig:convergence} shows the convergence plots of neural models on two datasets. We notice that other than the disaggregation from PUMA to TRACT on NYC Bikeshare dataset, SARN models always start with, and maintain lower validation error throughout training stage, thus converging faster than other models. One explanation are --- 1) given the same recurrent temporal module, SARN converges faster than GRU, indicating the power of spatial transformer layer. 2) The spatial attention layers in SARN models are aware of the spatial structures due to the containment maps, which provide "hints" during training, steering the learning toward the correct solutions (which always exhibit the containment property), thereby converging faster. 

\paragraph{\textbf{SARN Yields Better Disaggregation Results (Spatial Investigation):}} 

\begin{figure}[ht]
    \centering
    \includegraphics[width=0.46\textwidth]{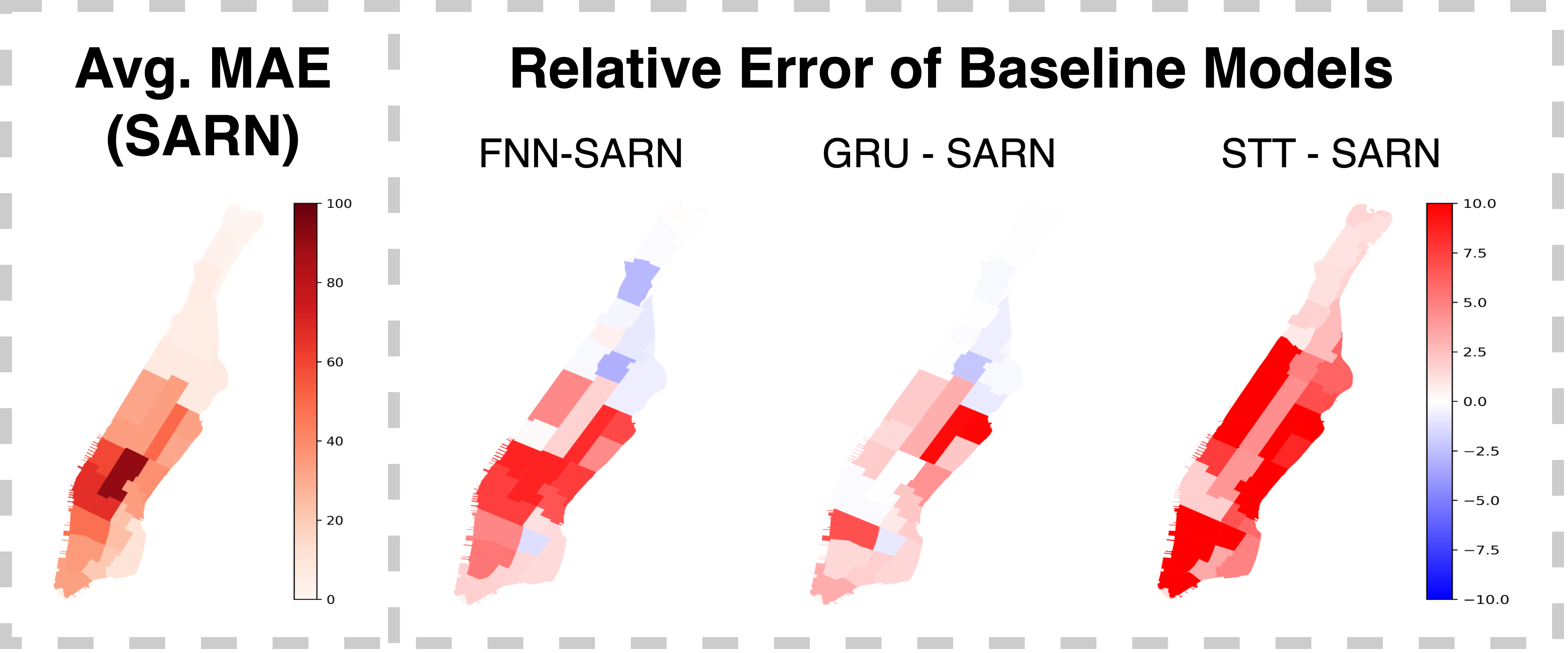}
    \vspace{-0.25cm}
    \caption{Left --- SARN tends to have higher errors in the central region of Manhattan. Right --- Overall, the three neural models produce larger errors (visualized as red regions). FNN and GRU generally have more errors in the central regions but perform similarly or slightly better in the upper regions. STT makes more errors across all regions.}
    \label{fig:relative-error}
    \vspace{-0.25cm}
\end{figure}

To facilitate understanding of the error distribution, we disaggregated NYC taxi data from PUMA to NTA using four neural models. Analyzing TRACT and BLOCK levels is challenging due to the large number of regions involved (283 and 3733 respectively). MAE error is calculated for all test samples and averaged for each NTA region. To compare performance, we calculate the difference in MAE error of each region between the other neural models and SARN, referred to as relative error. A positive relative error indicates that the neural model performs worse than SARN, while a negative relative error indicates better performance. Figure \ref{fig:relative-error} visualizes these errors. We observe the following: 1). SARN tends to have higher errors in the central region of Manhattan. This is reasonable given the higher traffic volume in the city center, making accurate disaggregation more challenging. 2) Overall, the other three neural models produce larger errors compared to SARN (visualized as red regions). FNN and GRU generally have more errors in the central regions but perform similarly or slightly better in the upper regions. STT, however, makes more errors across all regions. In conclusion, SARN consistently disaggregates data more accurately than the other neural models.

\begin{table}[ht]
    \small
    \tabcolsep=0.32cm
    \renewcommand*{\arraystretch}{1}
    \centering
    \begin{tabular}{|c|c|c|c|c|}
        \hline
        \textbf{Resolutions} & \textbf{Model} & \textbf{SARN} &\textbf{G.T.} &\textbf{\% Diff}\\ \hline
        \multirow{3}{*}{NTA} & LSTM & 184.665 & 184.035 & 0.03 \\
        & AGCRN & 124.911 & 137.625 & -9.24 \\
        & STSGCN & 50.441 & 50.134 & 0.61 \\ \hline 
        \multirow{3}{*}{TRACT} & LSTM & 23.945 & 23.973 & -0.11 \\
        & AGCRN & 9.764 & 9.808 & -0.45 \\
        & STSGCN & 10.586 & 10.145 & 4.35 \\ \hline 
        \multirow{3}{*}{BLOCK} & LSTM & 2.328 & 2.393 & -2.72\\
        & AGCRN & 1.648 & 1.452 & 13.47 \\
        & STSGCN & - & - & - \\ \hline 
        \multicolumn{4}{|c|}{Average \% Diff} & 0.78 \\ \hline 
    \end{tabular}
    \caption{Downstream application results (MAE) with disaggregated values at different geographic resolutions. SARN column shows the results from models with the disaggregated data with SARN model, while G.T. columns shows the results from models with the ground truth value. $\% Diff = \frac{SARN-G.T.}{G.T.}$. Negative \% Diff means better prediction results with SARN. STSGCN results at the BLOCK level is not reported due to out-of-memory issue.}
    \label{tab:application}
    \vspace{-0.5cm}
\end{table}

\subsection{Disaggregated Data is Authentic and Benefits Downstream Application}\label{sec:result-application}
Table \ref{tab:application} presents the next-hour traffic demand prediction results in MAE using disaggregated data from SARN at three geographic levels. These results are compared to those from models trained with ground truth data. We observe that, all models trained with disaggregated data exhibit performance that is very close to those trained with ground truth data, with minor and sometimes negligible differences. In some cases, the models perform even better with disaggregated data (e.g., NTA with AGCRN; TRACT with AGCRN; BLOCK with LSTM). Overall, the percentage difference is only 0.78\% on average, which is very trivial. These findings indicate that the disaggregated values produced by the SARN model are authentic, of high quality, and closely mimic the distribution of the ground truth data. This has significant real-world implications. Disintegrating aggregated data into finer resolutions and then forecasting traffic demand can capture hotspots and dynamic human mobility patterns that coarse, aggregated data cannot provide.

\subsection{Transfer Learning is Remarkably Effective}\label{sec:result-transfer-learning}
\begin{figure}[ht]
    \centering
    \vspace{-0.25cm}
    \includegraphics[width=0.46\textwidth]{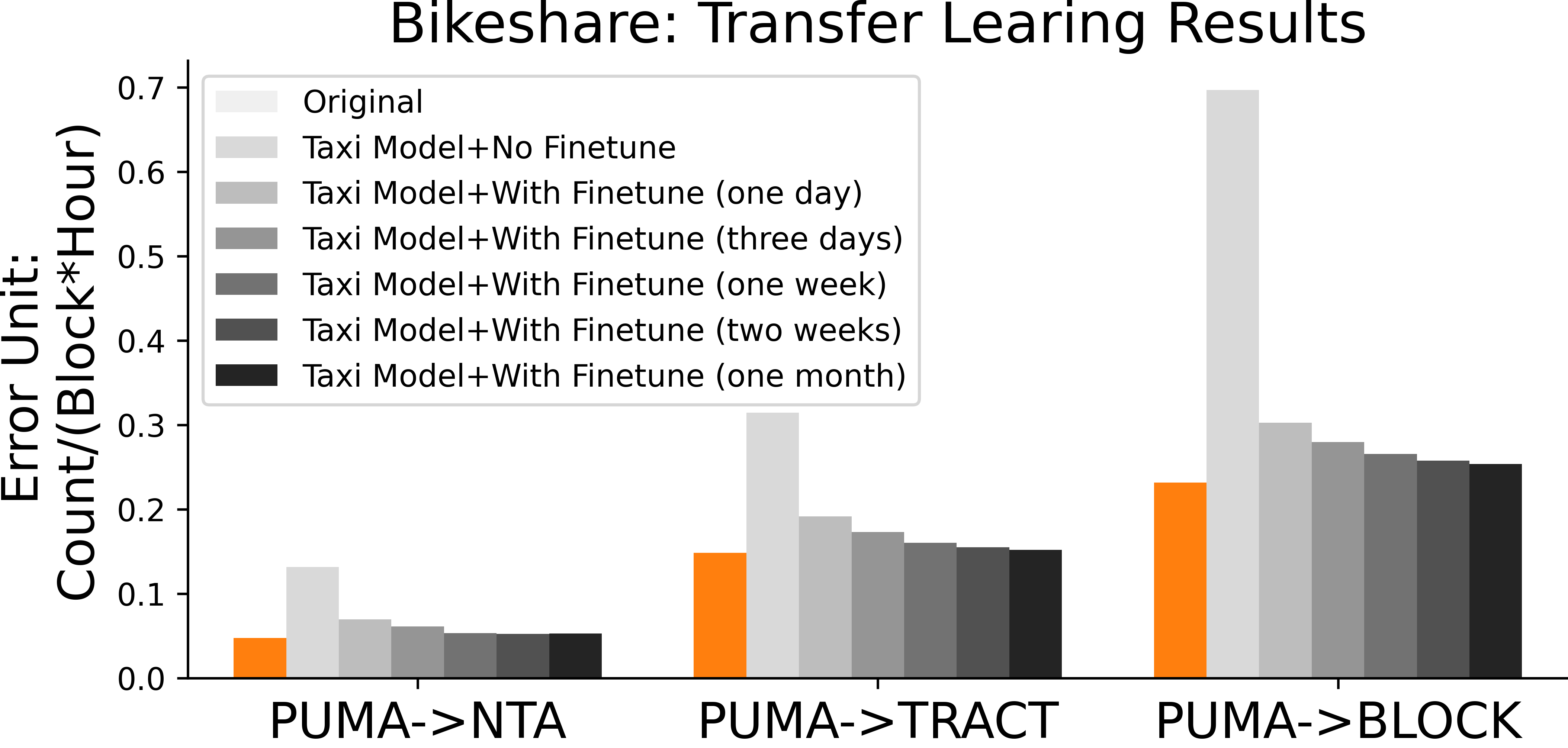}
    \vspace{-0.25cm}
    \caption{Transfer learning results on bikeshare dataset. The models are pre-trained with taxi data. The orange bar indicates the best performance for each disaggregation task.}
    \vspace{-0.25cm}
    \label{fig:transfer-learning}
\end{figure}

We trained SARN models on NYC taxi data and evaluated on bikeshare in the same NYC domain. We fine-tuned with one day (24 samples), 3 days (72), one week (168), two weeks (336), and one month of data (720). Figure \ref{fig:transfer-learning} shows the results: adding more fine-tuning samples enhances disaggregation performance. Surprisingly, fine-tuning with only one month of data achieves results comparable with the specialized model trained from scratch with bikeshare data. Additionally, even the performance slightly dropped, the benefit gain in saving computational costs is significant --- fine-tuning a pre-trained model greatly reduces the time costs (1$\sim$2 hours v.s. 12 hours). This result has significant implications for practical settings: Training data is not uniformly available for all variables due to technical, privacy, and political effects. But because cities are complex systems, all variables are interdependent~\cite{integrativeUrbanAI}. This result shows that pre-training can capture city dynamics resulting from the built environment and human activity, and that those signals tend to be predictive for multiple variables. We envision developing a core model pre-trained with ample historical records, then fine-tuning for different tasks in the same city, analogous to a language model being fine-tuned for more specialized tasks.

\subsection{Realistic Synthesis of Individual Events}\label{sec:result-point-synthesis}
\begin{figure}[ht]
    \centering
    \includegraphics[width=0.46\textwidth]{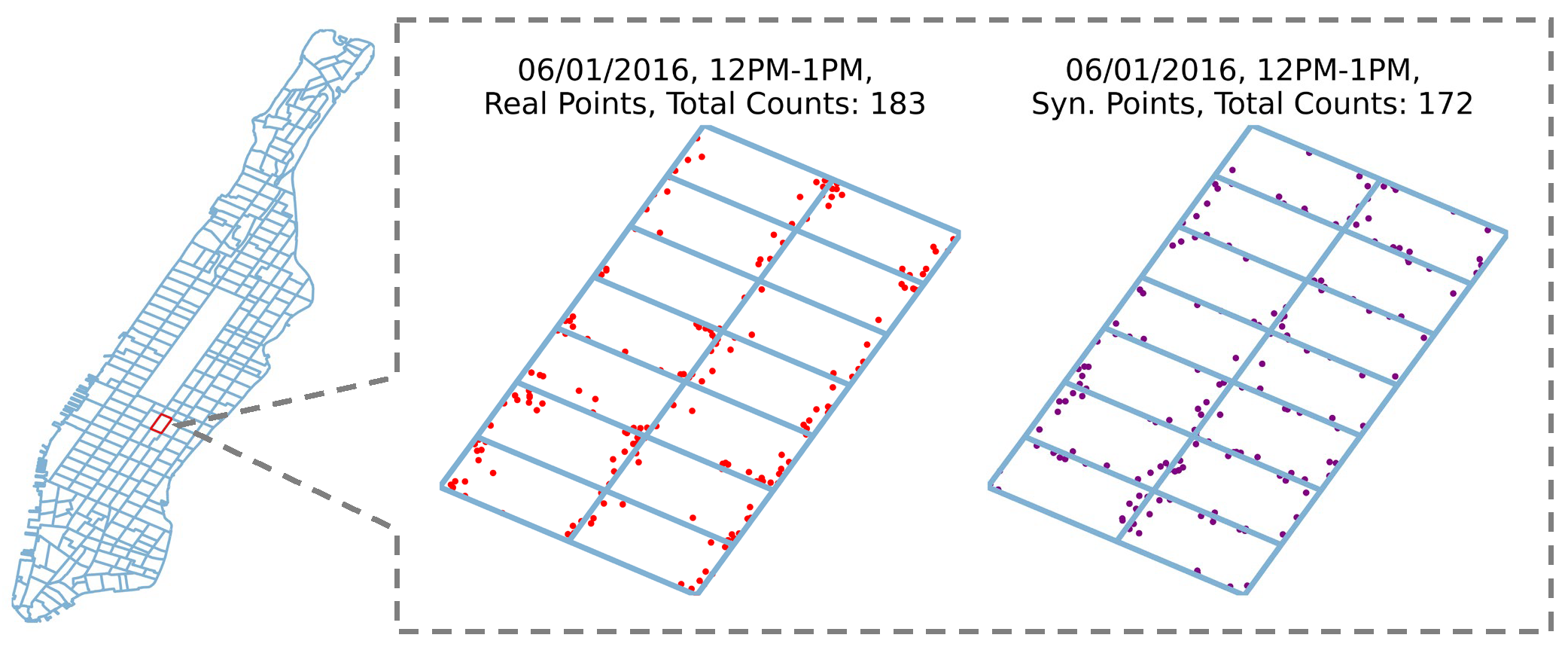}
    \caption{Synthetic individual-level taxi points. (Left) Randomly selected single tract region, containing multiple block regions. (Right) The distributions of real and synthesized taxis ride over the block regions are qualitatively similar.}
    \label{fig:point-distributions}
    \vspace{-0.25cm}
\end{figure}

To generate individual point data, we use SARN to disaggregate from PUMA to BLOCK with NYC taxi data. Then, we measure the mutual information and KL divergence (using Python implementations\footnote{\url{https://scikit-learn.org/stable/modules/generated/sklearn.metrics.normalized_mutual_info_score.html}}\footnote{\url{https://docs.scipy.org/doc/scipy/reference/generated/scipy.stats.kstest.html}}) between the distribution of disaggregated values and the original distribution in the same time period and region set. Then, for each BLOCK region with $k$ taxi counts, we randomly draw $k$ points within the region. We work with a set of block regions in Manhattan as shown in Figure \ref{fig:point-distributions}. We draw samples from this set of regions for the on 12PM-1PM, 06/01/2016.

The normalized MI scores is \textbf{0.7360}, smaller than 0.5, indicating that two distributions share more than half their information. The KL-divergence is \textbf{0.0582}. The low KL divergence indicates a reasonable approximation of the ground truth distribution using our disaggregated values. Qualitatively, Figure \ref{fig:point-distributions} depicts the distributions of both real and synthesized taxi location points within the region set from 12PM to 1PM. The model generated slightly smaller numbers of taxi rides compared to the real counts, with differences of 11 rides. While these differences are present, they are not significant compared with all taxi counts within the whole regions. In terms of the spatial distributions, the synthesized points capture the major clusters observed in the real taxi rides for both time periods, such as the clusters in the bottom left and center. This suggests that the synthesized points closely approximate the general patterns and trends present in the real data. On the other hand, the presence of discrepancies between the real and synthesized data points can have benefits, particularly in terms of privacy protection. As highlighted in the introduction, individual privacy is a significant motivator for \textit{not} releasing individual-level data. A closely matching distribution could be used as the basis of a privacy attack. For instance, clusters within very small areas may reveal the address of frequent riders. We consider our methods as providing better control over the trade-offs between faithful distribution and privacy (though without formal privacy guarantees, which are atypical for transportation data anyway).

\section{Discussion and Conclusions}\label{sec:discussion}
In this work, we synthesize high-resolution urban data by disaggregating low-resolution data.  We considered neural models relative to popular heuristic methods in the literature. FNN and CNN models were able to capture spatial connections among different regions, while GRU models further incorporated temporal interdependence and produced better results.

We proposed a structurally-aware recurrent network SARN model, which replaces fully-connected layers with structurally-aware spatial attention (SASA) layers in the GRU model. The spatial attention layers model the complex spatial interactions among regions, while the gated recurrent module captures temporal dependencies. Each SASA layer calculates global attention that captures broad spatial dependencies. It also calculates structural attention using the containment maps, ensuring a coherent and consistent disaggregation across different geographic resolutions. Our experimental results on both taxi and bikeshare dataset consistently demonstrated the effectiveness of our proposed SARN model, with statistically significant enhancements observed in almost all disaggreagtion tasks. It also converges faster than other neural models.

We conducted three supplementary experiments to demonstrate the utility of the disaggregated data --- 1) We use the disaggregated data in traffic demand prediction task, and show that the disaggregated data closely approximately the ground truth data with trivial performance difference, indicating its great quality and potential applicability in downstrea applications where coarse, aggregated data is less useful. 2) Since not all variables are available with significant training data, we considered the transferability of a model trained on one variable for predictions on another (in the same city). Surprisingly, with only a month of fine-tuning bikeshare data, the model exhibited performance comparable to that of the specialized model. This result exposes that the underlying dynamics of human activity may appear in all variables observed from the same built environment, motivating new applications that integrate multiple data sources to overcome limited data availability~\cite{khatiwada2022}. 3) We show that a simple sampling approach can synthesize individual events from aggregate data, exposing new opportunities for research.

There are some limitations to the study that we plan to address in future research: 1) we only considered the geographic levels for which  containment holds. However, our spatial attention calculation is not limited by the containment property. We can still construct the containment maps even when a finer region overlap with more than one coarse regions. However, the extent to which these containment maps enhance learning effectiveness remains to be investigated in future studies. 2) Our current transfer learning experiments exclusively focus on scenarios where datasets originate from the same city and domain. This assumption implies shared geographic structures/grids and similar underlying dynamics. However, in practice, A) two variables could stem from different domains within the same city, resulting in varied underlying dynamics despite sharing the same city structure; B) transfer learning across cities, where the underlying dynamics may be different, is an important use case that we do not consider, and 3) We only investigated one set of regions to synthesize individual points for one time period. The spatial and temporal variances might impact the quality of the synthesized points. Additionally, we did not investigate whether the ability to synthesize individual events introduces privacy risk. 4) Given the sequential processing nature of GRU model, SARN does incur heavy computational costs compared to other baseline models. For example, when disaggregating from PUMA to BLOCK with NYC datasets, it could take 12 hours of training, while FNN or GRU only a few hours. The computational burden is especially heavy when the disaggregated dimension is high (e.g., block level). We consider this as a limitation of using sequential models.

\clearpage
\bibliographystyle{ACM-Reference-Format}
\bibliography{disaggregation}

\end{document}